\documentclass[sigconf]{acmart}

\newcommand{\ie}{\emph{i.e.}, }

\newcommand{\eg}{\emph{e.g.}, }

\newenvironment{packeditemize}{\begin{list}{$\bullet$}{\setlength{\itemsep}{0.2pt}\addtolength{\labelwidth}{-4pt}\setlength{\leftmargin}{\labelwidth}\setlength{\listparindent}{\parindent}\setlength{\parsep}{1pt}\setlength{\topsep}{0pt}}}{\end{list}}

\usepackage{multirow}
\usepackage{caption}
\usepackage{subcaption}
\usepackage{makecell}

\AtBeginDocument{%
  \providecommand\BibTeX{{%
    \normalfont B\kern-0.5em{\scshape i\kern-0.25em b}\kern-0.8em\TeX}}}

\setcopyright{acmcopyright}
\copyrightyear{2021}
\acmYear{2021}
\setcopyright{acmlicensed}
\acmConference[CIKM '21] {Proceedings of the 30th ACM International Conference on Information and Knowledge Management}{November 1--5, 2021}{Virtual Event, Australia.}
\acmBooktitle{Proceedings of the 30th ACM Int'l Conf. on Information and Knowledge Management (CIKM '21), November 1--5, 2021, Virtual Event, Australia}
\acmPrice{15.00}
\acmISBN{978-1-4503-8446-9/21/11}
\acmDOI{10.1145/3459637.3481908}

\begin{document}
\fancyhead{}

\setlength{\abovedisplayskip}{2pt}
\setlength{\belowdisplayskip}{2pt}

\title[Detection of Illicit Drug Trafficking Events on Instagram]
{Detection of Illicit Drug Trafficking Events on Instagram: \\  A Deep Multimodal Multilabel Learning Approach}

\author{Chuanbo Hu, Minglei Yin, Bin Liu, Xin Li}
\authornote{This work is partially supported by the NSF under grants IIS-2107172, IIS-2140785, CNS-1940859, CNS-1814825, IIS-2027127, IIS-2040144, IIS-1951504 and OAC-1940855, the NIJ 2018-75-CX-0032. }

\affiliation{%
  \institution{West Virginia University}
  \city{Morgantown}
  \state{WV}
  \country{USA}
}
\email{cbhu@whu.edu.cn, my0033@mix.wvu.edu}
\email{bin.liu1@mail.wvu.edu, xin.li@mail.wvu.edu}

\author{Yanfang Ye}
\affiliation{%
  \institution{Case Western Reserve University}
  \institution{University of Notre Dame}
  \streetaddress{Dept. of CDS}
  \country{Ohio/Indiana, USA}}
\email{yanfang.ye@case.edu}

\renewcommand{\shortauthors}{Hu and Yin, et al.}

\begin{abstract}
Social media such as Instagram and Twitter have become important platforms for marketing and selling illicit drugs. 
Detection of online illicit drug trafficking has become critical to combat the online trade of illicit drugs. 
However, the legal status often varies spatially and temporally; even for the same drug, federal and state legislation can have different regulations about its legality. Meanwhile, more drug trafficking events are disguised as a novel form of advertising - {\em commenting} leading to information heterogeneity. Accordingly, accurate detection of illicit drug trafficking events (IDTEs) from social media has become even more challenging.
In this work, we conduct the first systematic study on {\em fine-grained} detection of  IDTEs on Instagram. 
We propose to take a deep  multimodal  multilabel learning (DMML) approach to detect  IDTEs and demonstrate its effectiveness on a newly constructed dataset called multimodal IDTE (MM-IDTE). Specifically, our model takes text and image data as the input and combines multimodal information to predict multiple labels of illicit drugs. Inspired by the success of BERT, we have developed a self-supervised multimodal bidirectional transformer by jointly fine-tuning pretrained text and image encoders. We have constructed a large-scale dataset MM-IDTE with manually annotated multiple drug labels to support fine-grained detection of illicit drugs. 
Extensive experimental results on the MM-IDTE dataset show that the proposed DMML methodology can accurately detect IDTEs even in the presence of special characters and style changes attempting to evade detection. 
\end{abstract}

\ccsdesc[500]{Information systems~Data mining}
\ccsdesc[300]{Applied computing~Law, social and behavioral sciences}

\keywords{drug trafficking; event detection; Instagram; multimodel bidirectional transformer; multilabel learning}

\maketitle

\def \Pr {{\mathrm{Pr}}}

\def \xx {{\bf x}}
\def \XX {{\bf X}}

\def \YY {{\bf Y}}
\def \yy {{\bf y}}

\def \ZZ {{\bf Z}}
\def \zz {{\bf z}}

\def \TT {{\bf T}}
\def \tt {{\bf t}}

\def \ww {{\bf w}}
\def \WW {{\bf W}}
\def \SS {{\bf S}}

\def \uu {{\bf u}}
\def \UU {{\bf U}}

\def \vv {{\bf v}}
\def \VV {{\bf V}}

\def \uww { \underline{\bf w}}
\def \oww { \overline{\bf w}}

\def \RR {{\bf R}}

\def \II {{\bf I}}
 \vspace{-10pt}
\section{Introduction}

\begin{figure*}[h]
  \centering
\begin{subfigure}[b]{0.46\textwidth}
\includegraphics[width=0.95\textwidth,height=2.0in]{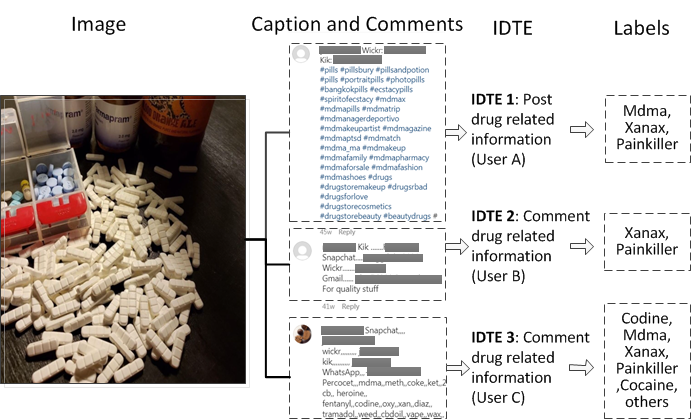}
  \subcaption{Direct drug advertising.}
  \label{fig:IDTE-direct}
\end{subfigure}
\begin{subfigure}[b]{0.5\textwidth}
\includegraphics[width=1.0\textwidth,height=2.0in]{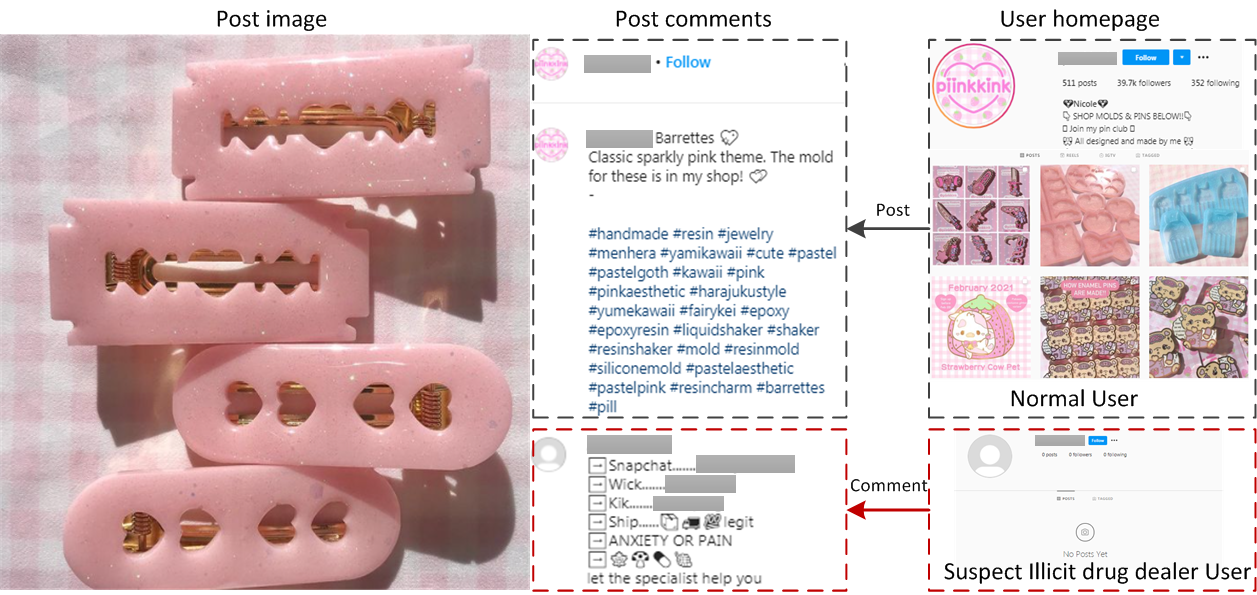}
  \subcaption{Indirect drug advertising.}
  \label{fig:IDTE-indirect}

\end{subfigure}

\vspace{-10pt}
 \caption{Examples of suspect IDTEs. Left: illicit drug dealers directly posting some drug-related texts or images. Right: illicit drug dealers indirectly advertise their products by adding a comment to an existing innocent post.}
 \label{fig:IDTE}
 \vspace{-10pt}
\end{figure*}

The co-evolution of cyberspace and human society has transformed the practice of illicit drug trade from the physical world to online platforms. Recent studies \cite{yang2017tracking,li2019machine,zhao2020computational,hassanpour2019identifying,kalyanam2017review,sarker2020mining,sarker2019machine} have shown that major social media platforms, including Instagram, Twitter, and Facebook, have become a  direct-to-consumer marketing tool for illegal drug dealers. 

What makes the combat on online illicit drug trade even more  challenging lies in the varying legal status of drugs. For instance, the use and possession of cannabis is illegal under federal law for any purpose in the US, but at the state level, policies regarding the medical and recreational use of cannabis vary greatly \cite{caulkins2016marijuana}. As of today, the recreational use of cannabis has been legalized in 15 states and decriminalized in another 16 states. Similarly, the legal status of club drugs varies according to the region and drug too. For example, some club drugs (e.g. cocaine) are almost always illegal; other club drugs (e.g.  amphetamine or MDMA) are generally illegal unless with a lawful prescription from a doctor; other drugs (e.g. "poppers") are legal in some jurisdictions.

Detection of online illicit drug trafficking becomes a critical step to combat the illicit online trade of illicit drugs. However, it is challenging  for the following reasons: (1) {\bf Inconsistency of drug legislation.} Legal status of drugs varies according to the region and drug. Most recently, Oregon's Measure 110 has decriminalized personal possession of small amounts of illegal drugs, such as cocaine, heroin, oxycodone, and meth; but the neighboring state of Washington has not. Such variation of legal status from state to state makes it difficult to draw a clear boundary between legal and illicit drug trade.
(2) {\bf Information heterogeneity.} The data sources related to drug trade involve both images and texts; meanwhile, the ways of advertising illicit drugs range from direct posting to indirect commenting (as a disguised form of advertising). More importantly, there is additional uncertainty arising from  illicit drug dealer's attempting to evade detection by different means (e.g., changing font styles, adding separators between letters, coining new street names of popular  drugs). How to systematically combine these heterogeneous information has remained open.
(3) {\bf Accuracy.} When compared with normal users posting images and texts related to legal drug use, the portion of illicit drug trafficking is relatively small. Searching for illicit drug trafficking activities is like finding a needle in a haystack. How to achieve a low false alarm rate while efficiently mining a large amount of social media data calls for innovative technical solutions at the system level. 

In this paper, we propose a deep  multimodal  multilabel learning (DMML) approach to detect the existence of  multiple illicit drugs from suspect illicit drug trafficking events (IDTEs) on Instagram. As shown in Figure \ref{fig:IDTE}, a suspect IDTE is a user activity such as a post or a comment following a post on Instagram.  Note that a suspect IDTE can be an initial post meant for  marketing of illegal drugs. It can also be a comment following an initial post, in which drug trafficking information is added  even though the original post does not contain any drug information. 
Unlike existing works on drug dealer detection \cite{yang2017tracking,li2019machine,zhao2020computational} or drug use detection \cite{hassanpour2019identifying,kalyanam2017review,sarker2020mining,sarker2019machine} from aggregated information, we advocate to focus on detecting activities related to suspect IDTEs. This is because as the arms race between drug dealers and law enforcement evolves, more drug trafficking events are disguised as a novel form of advertising - {\em commenting}. Instead of directly posting some drug-related text or images (easily caught by the regulation), illicit drug dealers often indirectly advertise their products by adding a comment on the existing harmless post (refer to Fig. \ref{fig:IDTE-indirect}). Note that such a piggyback strategy can be recursively applied, so the event of drug trafficking (drug-related comments) can be embedded at several levels under the original post. Our work is also different from existing works \cite{yang2017tracking,li2019machine,hassanpour2019identifying,zhao2020computational} in the sense that our approach detects not only illicit drugs but also their specific types in each suspect IDTE. Such {\em fine-grained} detection of illicit drug trafficking becomes particularly important considering the inconsistency of drug legislation across different states.

Specifically, our model takes in text and image data associated with suspect IDTEs and composites the multimodal information to predict multiple labels of an illicit drug. Motivated by the latest advances in natural language processing - e.g., Bidirectional Encoder Representations from Transformers (BERT) \cite{devlin2018bert}, Vision-and-Language BERT (ViLBERT) \cite{lu2019vilbert}, Learning Cross-Modality Encoder Representations from Transformers (LXMERT) \cite{tan2019lxmert}), we propose to develop a self-supervised multimodal  bidirectional transformer (MMBT) by jointly fine-tuning pretrained text and image encoders. By projecting image embeddings to the text token space, we can employ self-attention over both
modalities simultaneously, achieving more fine-grained multimodal fusion \cite{kiela2019supervised}. The combined multimodal feature is then passed to a multi-label learning module for predicting the type of multiple illicit drugs.

We have manually constructed a large-scale multimodal IDTE (MM-IDTE) dataset for the purpose of fine-grained illicit drug detection. Our MM-IDTE dataset, containing nearly 4,000 posts and more than 6,000 comments, represents the largest multimodal (text+image) illicit drug detection dataset so far. In particular, to construct such a large-scale dataset, we have designed an automatic data crawling system for Instagram that jointly uses hashtag and image information to guide the data collection. We have spent hundreds of hours on manually annotating each post (text and images) by multiple labels. Such multilabel ground truth has been verified by different people to ensure their consistency and accuracy. Extensive experimental results on the constructed dataset show that the proposed MMBT-based DMML approach can accurately detect IDTE. Both  micro-/macro- precision and recall performance of our approach exceeds 0.90 on the test MM-IDTE dataset.

The key contributions of this paper are summarized as follows. 
\begin{packeditemize}
    \item We conduct the first systematic study on fine-grained detection of illicit drug trafficking events on Instagram. Unlike existing works on drug dealer or abuse detection, this work focuses on addressing the issue of commenting as a disguised form of advertising on Instagram.
    
    \item We propose a deep multimodal  multilabel learning (DMML) framework for detecting illicit drug trafficking events. It is shown that the proposed MMBT-based approach can dramatically outperform unimodality and ad hoc multimodal fusion strategies. We have also experimentally compared different image encoders for MMBT-based fusion and found that MMBT based on ResNet50 and BERT achieves the best performance. 
    
    \item We construct a large-scale MM-IDTE dataset for fine-grained  illicit  drug  detection. Toward this objective, we have developed an automatic hashtag-based data crawling system and a user-friendly data annotation system to support large-scale and multimodal data collection. The newly constructed MM-IDTE dataset will be made publicly available to support the research related to illicit drug trafficking activities.
    
    \item We demonstrate the effectiveness of the proposed MMBT method on the MM-IDTE dataset. It is found that our method can successfully identify some challenging cases difficult for untrained eyes (e.g., special symbols and style changes attempting to evade detection). The developed system  could facilitate the disruption of illicit drug trade by law enforcement.
\end{packeditemize}

 \vspace{-10pt}
\section{Related Work}
\label{sec:relate}

\subsection{Drug Abuse and Dealing Analysis}
As far as we know, there has been limited work on tracking drug abuse and illicit drug trades from online data. Among these existing works, \cite{buntain2015your} analyzed the time and
location patterns of drug use by mining Twitter data; network information of Instagram user timelines was used in \cite{correia2016monitoring} to monitor suspicious drug interaction activities; \cite{zhou2016understanding} and \cite{yang2017tracking} analyzed Instagram data for tracking and identifying drug dealer accounts. More recently, machine learning and natural language processing techniques have been applied to combat prescription drug abuse \cite{kalyanam2017review,sarker2020mining,sarker2019machine,hassanpour2019identifying} and detect illicit drug dealers \cite{yang2017tracking,li2019machine,zhao2020computational}.

Our work is different from the previous works. First, our work focuses on suspected illicit drug trafficking events, while previous work focused on drug abuse or dealing with mining  from aggregated information. Second,  our method detects all potential illicit drugs in each suspected illicit drug trafficking event, while previous work either identified drug dealers or detected existing drug abuse. Third, we target at a fine-grained detection of different drugs as well as drug-related activities. Technically, we formulate our work as a multilabel learning problem \cite{zhang2013review}, which is much more challenging than the binary classification in the previous works. 

\subsection{Multimodal Learning and Data Fusion}

In many real-world problems, objects always involve multiple modalities.  A modality refers to the way in which an object is represented. The goal of multimodal learning is to design a strategy to leverage the information from multiple modalities so that different sources of information can complement and enhance each other for a specific goal~\cite{baltruvsaitis2018multimodal}. It usually involves a joint representations of different modalities and a way to fuse the representations to a composite multi-modal feature for the sake of the task in investigation. Multi-modal learning has enabled a wide range of applications such as multimedia content indexing and retrieval \cite{chang2003cbsa}, image captioning \cite{yu2019multimodal}, and visual question answering (VQA) \cite{fukui2016multimodal}. 

Rapid advances of machine learning in recent years have also expedited the research in multimodal data fusion \cite{zhang2020multimodal}. A straight way is to concatenate features from different modalities \cite{kiela2018efficient}. Bilinear pooling \cite{fukui2016multimodal} based method was proposed to better capture the interactions between features in different modalities. 
A gated multimodal fusion module was proposed in \cite{arevalo2017gated} to find an
intermediate representation based on a combination of data from different modalities. Its follow-up work has shown that fusion with discretized features outperforms text-only classification \cite{kiela2018efficient}. 
More recently, inspired by the success of Bidirectional Encoder Representations from Transformers (BERT) \cite{devlin2018bert}), transformer-based multimodal data fusion has attracted increasingly more attention - e.g., multimodal bitransformer
(MMBT) \cite{kiela2019supervised}, Vision-and-Language BERT (ViLBERT) \cite{lu2019vilbert}, Learning Cross-Modality Encoder Representations from Transformers (LXMERT) \cite{tan2019lxmert}.

\subsection{Multi-label Learning}

Multi-label learning \cite{zhang2013review} targets at representing an object by a single instance but each object can be associated with a set of labels. In contrast to traditional supervised learning, the task of multilabel learning is to learn a function that can predict the proper label sets for unseen instances. 
Typically, the multilabel learning problem is transferred into other well-established learning settings such as binary classification, one-vs-all classification, or multiclass classification through the introduction of label powerset~\cite{tsoumakas2010random}. 
There are some efforts to adopt learning techniques such as low-dimensional label embedding methods~\cite{bhatia2015sparse}, joint global and local approach \cite{zhu2017multi}, and joint learning of label-specific features and label correlations \cite{zhang2018multi},  to exploiting label correlations for multi-label learning. 

Deep multi-label learning has also been recently studied for image classification \cite{song2018deep} and in the special situation of extreme multi-label learning (XML) \cite{XMLCNN:sigir17,you2019attentionxml}.

\section{Illicit Drug Trafficking Event Detection}
\label{sec:model}

In this section, we first formulate the problem of \emph{illicit drug trafficking event detection} (IDTE), and then introduce the proposed \emph{deep multimodal multilabel learning} (DMML) approach.

\subsection{Problem Formulation}

\begin{definition}[Illicit drug trafficking event]
    An illicit drug trafficking event (IDTE) is an event on Instagram that contains the  marketing and selling of one or more defined illicit drugs. 
\end{definition}
In this paper, we consider the following nine common illicit drugs traded on Instagram: Marijuana, Codeine, Mdma, Xanax, Painkillers, Mushrooms, LSD, Cocaine, and other drugs. 

\begin{definition}[Suspect IDTE]
    A suspect IDTE is a user activity such as a post or a comment to a post on Instagram. It usually contains image and text information. 
\end{definition}
Figure \ref{fig:IDTE} shows some examples of suspect IDTEs on Instagram. Note that a suspect IDTE can be an initial post (\eg the post initialized by User A in Fig. \ref{fig:IDTE-direct}). It can also be  comments following a post, in which drug trafficking information is added (\eg comments by Users B and C in Fig. \ref{fig:IDTE-direct}). As shown in Fig. \ref{fig:IDTE-indirect}, even the initial post does not include any drug-related information, illicit drug dealers can still  advertise their products by adding a comment to an existing innocent post. 

Given the above definitions, we can formally define the problem of \emph{illicit drug trafficking event detection} as follows: The goal is to build an effective approach to detect the \emph{existence} of illicit drugs such as cocaine and cannabis within each suspect IDTE $i$. Assume there are a total of $C$ predefined illicit drugs under consideration. Let $y_{ic}=1, c\in\{1,\dots, C\}$ denotes suspect IDTE $i$ contains drug $c$ and $y_{ic}=0$ otherwise. Note that, it is also possible that the suspect IDTE $i$ does include any of the $C$ illicit drugs. We add one more label $y_{i0}= 1$ to indicate the case when the suspect IDTE is {\em drug-free}. Then each suspect IDTE $i$ can be represented as a vector $\yy_i=[y_{i0}, y_{i1}, y_{i2}, \dots, y_{iC}]\in \{0, 1\}^{C+1}$. Each suspect IDTE $i$ is associated with  an image $x_i$, and a text comment $t_i$, which has $T$ tokens. Given a set of $N$ training instances $\mathcal D=\{([x_1,t_1], \yy_1), ([x_2,t_2], \yy_2), \dots, ([x_N,t_N], \yy_N)\}$, we aim to build a predictive model $f: [x, t]  \rightarrow  \yy $ to detect the \emph{existence} and {\em types} of illicit drugs from new suspect IDTEs. 

\begin{figure}[t]
  \centering
  \includegraphics[width=\linewidth]{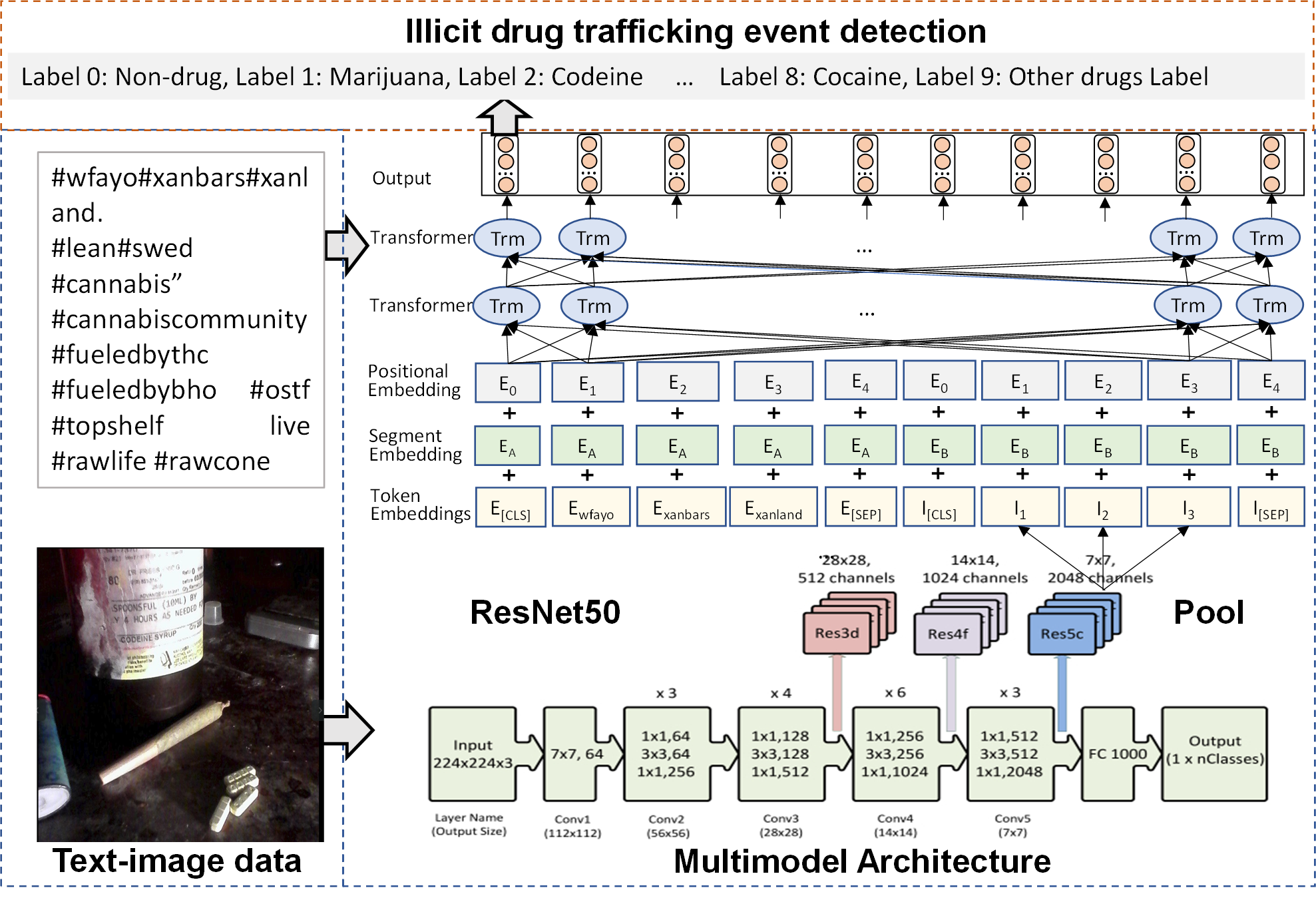}
  \vspace{-20pt}
  \caption{Overview of our proposed deep multi-modal multi-label learning approach to illicit drug trafficking activity detection. Our model takes in text and image data associated with suspect IDTEs, and combines multimodal information to predict multiple labels of illicit drugs.}
  \Description{}
  \label{fig:overview}
  \vspace{-20pt}
\end{figure}

\subsection{System Overview}
Figure \ref{fig:overview} shows the overview of our deep multimodal multilabel learning (DMML) approach to illicit drug trafficking event detection. The model takes in text and images associated with each suspect IDTE, and composites the text and image inputs via a multimodal bidirectional transformer unit. The composited text-image feature is then passed to a multi-label learning module to predict the  \emph{existence} and {\em types} of illicit drugs. 

It is worth mentioning several salient features of our DMML approach before elaborating on its details. First, the complementary role played by text and image information has been recognized by previous work (e.g., \cite{yang2017tracking}). However, it is unclear which modality contributes more to the detection especially when the problem of detection reaches fine-grained. Contrary to the findings reported in \cite{yang2017tracking}, we have found texts are more reliable than images for fine-grained drug classification.  Second, we target at extracting multilabel information from IDTE to more accurately track the spatio-temporal dynamical patterns of different drugs. Such feature is important to address the issue of inconsistency of drug legislation across different states.

\subsection{Multimodel Fusion via Bidirectional Transformer}

Each suspect IDTE  is associated with  an image $x$, and a text comment $t$, which has $T$ tokens. Let $t = (w_1, \dots, w_T)$.  The presence of multiple information sources holds the promise to learn better feature representations for final detection of illicit drugs.  Typically, text data $t$ is processed by a sequence model such as LSTM to form a text feature $\phi_t$, and image $x$ is processed using a pre-trained CNN model to form a image feature $\phi_x$. Then a fusion model is applied to combine the text and image features $\phi_{xt}=f_{\mathrm{fusion}}(\phi_x, \phi_t)$. There are a wide range of fusion methods proposed \cite{zhang2020multimodal} such as concatenation \cite{kiela2018efficient} and bilinear pooling \cite{gao2020revisiting,jia20203d}. 

In this paper, we propose to adapt a bidirectional transformers approach to multimodel fusion. First, we found textual are more reliable than images for fine-grained drug classification. Second, as discussed in \cite{wang2020makes},  multimodal models that composite features at late stages are prone to overfitting.  The bidirectional transformers approach projects image embedding to the text token space to form better feature representations for final detection of illicit drugs. 

\subsubsection{Transformers in Text and Vision}

The idea of self-supervised embedding or transfer learning from pre-trained representations has been extensively explored in the literature of natural language processing (e.g., \cite{mikolov2013distributed,kiros2015skip}) and computer vision (e.g., \cite{oquab2014learning,sharif2014cnn}). Most recently, the idea of fine-tuning self-supervised or semisupervised learning has revolutionalized the field of natural language processing leading to breakthroughs such as BERT \cite{devlin2018bert} and its variations (e.g., xlnet \cite{yang2019xlnet} and albert \cite{lan2019albert}). By applying the bidirectional training of transformer, a popular attention model, to language modelling, BERT learns contextual relations between words in a text more effectively.

The success of transformer architectures has rapidly leveraged to the field of computer vision \cite{han2020survey}. By integrating self-attention with self-supervision, transformers can exploit long-range dependencies in the input domain, which make transformer-based representation more expressive. Since there is a minimal assumption about prior knowledge, pretrained transformers are particularly suitable for large-scale and unlabelled datasets. Thanks to the generalization of encoded features, learned representations can be fine-tuned by labelled data leading to excellent performance on various vision tasks \cite{khan2021transformers}. 

\subsubsection{Multimodal Transformer}

Inspired by the latest advances in multimodal transformers (e.g., ViLBERT \cite{lu2019vilbert}, LXMERT \cite{tan2019lxmert}), we propose to develop a self-supervised multimodal bi-transformer (mimicking bidirectional transformers) jointly fine-tuning pretrained text and image encoders \cite{kiela2019supervised}. The basic idea is to use self-attention over both texts and images simultaneously, providing early and
fine-grained multimodal fusion. It has been shown in previous work \cite{kiela2019supervised} that such conceptually simple strategy can work as effectively as more sophisticated multimodally pretrained ViLBERT models. 

More specifically, we have constructed a multimodal bitransformer (MMBT) model (refer to Fig. \ref{fig:overview}) combining text-based self-supervised representations with image-based CNN architectures (e.g., ResNet \cite{he2016deep}). By projecting image embedding to the text token space, we can employ self-attention over both modalities simultaneously, achieving more fine-grained multimodal fusion \cite{kiela2019supervised}. It should be noted that a salient feature of MMBT is the flexibility of plugging in different images and text encoders. For example, the ResNet-152 image encoder adopted in \cite{kiela2019supervised} can be readily replaced by a smaller ResNet-50 counterpart.

\subsection{Multilabel Learning}

The output of our model is the predicted probability vector $\hat {\yy}_i =[\hat y_{i0}, \hat y_{i1},  \dots, \hat y_{iC}]\in \{0, 1\}^{C+1}$ for suspect IDTE $i$. This is a typical example of multilabel learning, where each example represented by a single instance is  simultaneously associated with multiple class labels \cite{zhang2018binary}. Under the context of IDTE detection, multilabel learning achieves fine-grained classification of multimodal data. To the best of our knowledge, such fine-granularity classification has not been considered in previous works of post-based drug dealer detection \cite{yang2017tracking}. 
Though there are different training strategies in multilabel learning \cite{zhu2017multi}, we have adopted the binary cross-entropy (BCE) loss, which has been widely applied in deep learning based multilabel learning problems~\cite{XMLCNN:sigir17,you2019attentionxml}.  The BCE loss is defined as
\begin{equation*}
    \ell(\Theta)=-\frac{1}{N}\sum_{i=1}^{N}\sum_{c=0}^{C} \left[ y_{ic}\log\hat y_{ic} + (1- y_{ic}) \log (1- \hat y_{ic}) \right].
\end{equation*}

\section{Multimodal IDTE (MM-IDTE) Dataset Construction}
\label{sec:data}

In this section, we present a new multimodel dataset constructed from Instagram; toward this objective, we will discuss our effort on data crawling and data annotation, respectively.

\begin{table}[th]
  \centering
  \small
  \caption{Comparison of our MM-IDTE dataset with several existing datasets for drug dealer and abuse  detection.}
  \vspace{-10pt}
  \resizebox{1.0\columnwidth}{!}{
    \begin{tabular}{lrlll}
    \hline
    \multicolumn{1}{l}{Study} & Source & \multicolumn{1}{l}{Granularity(\#)}  & classification & Application \\
    \hline
    \multicolumn{1}{r}{\cite{zhou2016fine}} & Instagram &  \makecell[l]{Post-level 2,362}  & Binary  & \makecell[l]{Drug use \\ pattern analysis} \\
    \hline
    \multicolumn{1}{r}{\cite{yang2017tracking}} & Instagram &  \makecell[l]{Post-level: 4,819}   & Binary  & \makecell[l]{Drug dealer \\ detection} \\
    \hline
    \multicolumn{1}{r}{\cite{mackey2018solution}} & Twitter & Tweets-level: 213,041  & Binary   & \makecell[l]{Drug dealer \\ detection} \\
    \hline
    \multicolumn{1}{r}{\cite{hu2019insight}} & Twitter & Tweets-level: 1,794 & Binary  & \makecell[l]{Drug abuse \\ risk detection} \\
    \hline
    \multicolumn{1}{r}{\cite{hassanpour2019identifying}} & Instagram  & Post-level: 369,000  & Binary & \makecell[l]{Drug abuse \\ risk detection}  \\
    \hline
    \multicolumn{1}{r}{\cite{li2019machine}} & Instagram &  \makecell[l]{Post-level:1,228}  & Binary   & \makecell[l]{Drug dealer \\ detection} \\
    \hline
    Ours  & Instagram &  \makecell[l]{IDTE-level: 4,648} & Multi-label & IDTE detection  \\
    \hline
    \end{tabular}%
    }
  \label{tab:datasets}%
  \vspace{-10pt}
\end{table}%

\begin{figure*}[!h]
  \centering
  \includegraphics[width=0.8\linewidth]{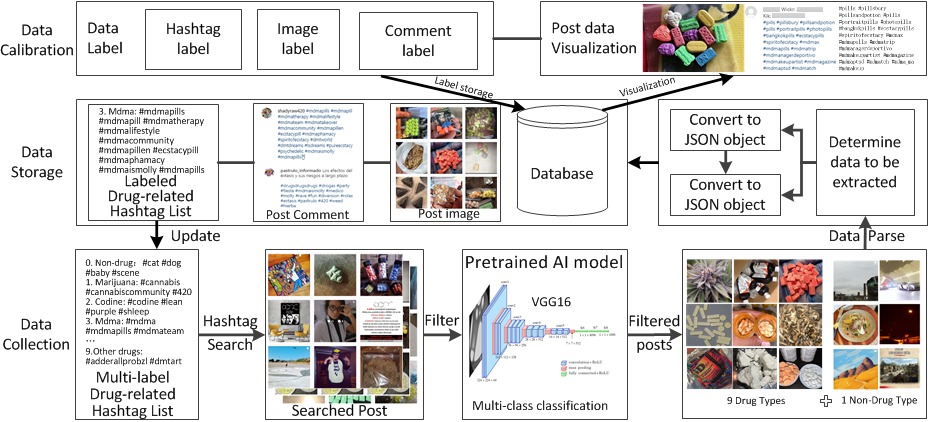}
  \vspace{-10pt}
  \caption{Data collection, storage, and calibration: note that our annotation involves multiple labels and multiple modalities, which requires hundreds of hours of manual labor for cross-checking and validation.}
  \Description{}
  \label{fig:data_collection_label}
  \vspace{-10pt}
\end{figure*}

To achieve this objective, we have made a great effort on data collection and construction of multimodel IDTE dataset in this project. An integrated data collection and calibration platform has been designed for automatic data collection/crawling and synchronized multilabel data calibration, as shown in Figure \ref{fig:data_collection_label}. Three main modules in the platform have been implemented to realize suspect IDTE data collection, data storage, and data calibration, respectively. 

\subsection{Data collection}

The rationale underlying our data crawling scheme is still based on {\it hashtag}-based search \cite{godin2013using}. Hashtags on Instagram can help users extend their reach, engage their audience, which can be attached to posts, and become clickable phrases and topics with the $\#$ placed in front of them. However, unlike \cite{yang2017tracking} working with a fixed collection of hashtags, we propose a data crawling algorithm that iteratively expands the pool of hashtags for scaling up our search. Such expansion of hashtags is guided by an intelligent pretrained AI model (VGG-16 \cite{simonyan2014very}) designed for drug image classification. By treating drug-related hashtags and images as a pair of peer hidden variables, our iterative crawling system aims at refining and updating the collected multi-modal data in an Expectation-Maximization (EM)-like manner. The detailed description of our data collection system consists of the following four components.

\begin{itemize}
\item[1.]  Drug-related hashtags collection: A total of 200 drug-related hashtags have been manually collected by domain experts using the hashtag search API \cite{gao2017hashtag}. These hashtags contain 10 types of drugs (i.e., non-drug, marijuana, codine, 3,4-methylenedioxy-methamphetamine (MDMA), xanax, painkiller, psilocybin mushroom (hereinafter called mushroom), Lysergic acid diethylamide (LSD), cocaine, other drugs), which are widely trafficked on Instagram\footnote{https://drugabuse.com/featured/instagram-drug-dealers/}. We have used this set of hashtags as the initial starting point of our data collection.
\item[2.]  Drug-related post detection. We search each post (which includes an image and comments) with each drug-related hashtag as input. A VGG-16 based binary classification model \cite{simonyan2014very} is pretrained to detect drug-related posts from the accompanying image information. The image-based dataset for model pretraining contains various types of drug-related images which are sources from Bing image search API (similar to Google image search API adopted in \cite{yang2017tracking}). If an image of a post is detected by the model as being drug-related (positive), we save its link for further processing.
\item[3.] Drug-related data collection. The detected posts were converted and formalized into a universal {\em json} object to facilitate the storage and retrieval. As post comments are sources from several user accounts, we saved each post-related information (including posted images and comments). Totally, 10,000 potential posts and 23,034 user homepage information were collected as the initial dataset. 
\item[4.] Drug-related hashtag update. New hashtags from each detected post can be added into the list of drug-related hashtags. We have also recorded the frequency of each hashtag in order to track the most frequent ones. The system uses the new hashtag (which have the highest frequent counts) in the next iteration until the amount of collected data reaches a prespecified threshold (in this study, we have set the threshold to be 1000 drug-dealer accounts). 
\end{itemize}

\subsection{Data annotation}
Suspect IDTEs often contain various types of drugs, so we have designed a multi-label annotation module, which contains Instagram post data visualization and three information (i.e., Hashtag label, image label and comment label) annotation. The three information of any suspect IDTEs can be labeled by domain expert users through 10 optional categories. The labeled hashtags will update the drug-related hashtag list weight to improve data collection efficiency. Totally, 4,648 suspect IDTEs were labeled during the construction of the experimental dataset, which contains 1,022 drug trafficking posts and 1,406 unique drug dealer user accounts. The proportion of each category label in the dataset is shown in Figure \ref{fig:data_dist}. Several existing datasets have been designed for illicit drug dealing tracking, as shown in Table \ref{tab:datasets}. Compared with these existing datasets, our dataset contains more user accounts and more diverse labels. 

To ensure the consistency and accuracy of multiple labels, we have asked different people to cross-validate the annotation results. Several particular challenges we have identified through the manual data annotation process include: 1) certain drugs (e.g., power-type) have similar visual appearance and can be easily confused with each other (e.g., DMT vs. MDMA); 2) the same type of type (\eg MDMA) can have different visual appearances (e.g., power vs. pills); 3) an image or a post can contain a large number of different drugs (easy to miss some). To help human annotators with the labeling process, we have designed a user-friendly interface consisting of data calibration at three different levels (hashtag label, comment label, and image label). It is estimated that a total of over 200 hours have been spent on manually annotating the collected dataset.

 \vspace{-5pt}
\section{Experimental Results}
\label{sec:exp}

\subsection{Experimental Setup}
\begin{figure}[t]
  \centering
  \small
  \includegraphics[width=0.95\linewidth]{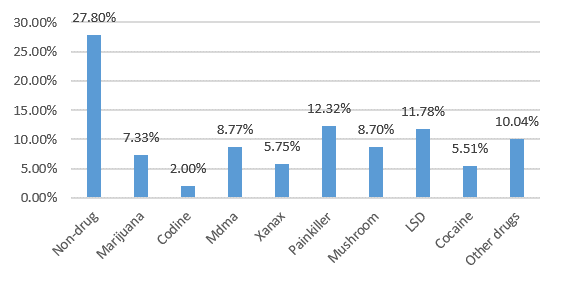}
  \vspace{-15pt}
  \caption{Distribution of labels in the IDTE dataset.}
  \Description{}
  \label{fig:data_dist}
  \vspace{-15pt}
\end{figure}

\noindent{\bf Data.} The MM-IDTE dataset used in our experiment contains 4,648 manually labeled IDTE records. There are 10 labels including 9 drug labels (\ie, Marijuana, Codeine, Mdma, Xanax, Painkillers,Mushrooms, LSD, Cocaine, and other drugs) and a non-drug label when the IDTE is drug-free. Figure \ref{fig:data_dist} shows the distributions of the 10 labels across the whole IDTE records. 

\noindent{\bf Training and testing.}  We randomly split the dataset into a training set (75\%) and a testing set (25\%). We have trained the models using the popular Adam optimization algorithm \cite{reddi2019convergence}.

The following parameters are adopted in our setting: learning rate $\alpha=2e^{-5}$, $\beta_1=0.9,\beta_2=0.999$, $\epsilon=10^{-8}$. We opt to terminate the training after 50 epochs. All experiments are conducted using PyTorch on a workstation with one RTX 2080 GPU.

\begin{table*}[th]
\begin{center}
\caption{Performance comparison between the proposed method  and the baseline approaches.}\label{tab:classfication}
\vspace{-10pt}
\begin{tabular}{ | c|c c|c c c| c c c|  }
 \hline
 & \multicolumn{2}{c |}{Example-based } & \multicolumn{6}{c |}{Label-based }  \\
 \hline
Method & Subset Accu. & Hamming loss & micro Pre & micro Recall & micro F1  & macro Pre & macro Recall & macro F1 \\ \hline
ResNet50 & 0.4187 & 0.3054 & 0.3348 & 0.7197 & 0.4570  & 0.2949 & 0.6635 & 0.3750  \\ 
ResNet152 & 0.4155 & 0.2844 & 0.2901 & 0.6612 & 0.4537  & 0.3453 & 0.5717 & 0.3500  \\ 
VGG16 & 0.4144 & 0.4066 & 0.2901 & 0.8499 & 0.4275  & 0.2855 & 0.8986 & 0.3862 \\ 
DenseNet121 & 0.4263 & 0.3551 & 0.2916 & 0.7559 & 0.4252  & 0.2790 & 0.7378 & 0.3576 \\ \hline
TextRNN & 0.6775 & 0.0831 & 0.7524 & 0.6848 & 0.7170  & 0.6829 & 0.4896 & 0.5462 \\ 
TextCNN & 0.6502 & 0.0956 & 0.6918 & 0.6826 & 0.6872  & 0.5853 & 0.5144 & 0.5374 \\ 
FastText & 0.6604 & 0.0875 & 0.7272 & 0.6892 & 0.7077  & 0.6892 & 0.4870 & 0.5369\\ 
BERT & 0.8902 & 0.0208 & 0.9174 & 0.9711 & 0.9435  & 0.8786 & 0.9671 & 0.9185\\ \hline
Concatenation & 0.5727 & 0.1255 & 0.5986 & 0.9024 & 0.7197  & 0.5333 & 0.8806 & 0.6395\\ 
FBC & 0.4553 & 0.2315 & 0.4261 & 0.8547 & 0.5687 & 0.3582 & 0.8547 & 0.4655\\  \hline
\bf Proposed & \bf 0.9322 & \bf 0.0135  & \bf 0.9496 & \bf 0.9765 & \bf 0.9629 & \bf 0.9217 & \bf 0.9728 & \bf 0.9455\\ \hline
\end{tabular}
\end{center}
\vspace{-10pt}
\end{table*}

\begin{table*}[th]
  \centering
  \caption{Impact of different image encoders on our method.}
  \vspace{-10pt}
    \begin{tabular}{|cc|rr|rrr|rrr|}
    \hline
          &       & \multicolumn{2}{|c|}{Example-based} & \multicolumn{6}{c|}{Label-based} \\
    \hline
    \multicolumn{2}{|c}{Method} & \multicolumn{1}{|l}{Subset Accu.} & \multicolumn{1}{l}{ Hamming loss} & \multicolumn{1}{|c}{micro Pre  } & \multicolumn{1}{c}{micro Recall} & \multicolumn{1}{c}{micro F1} & \multicolumn{1}{|c}{macro Pre  } & \multicolumn{1}{c}{macro Recall} & \multicolumn{1}{c|}{macro F1} \\
    \hline
    \multirow{7}[0]{*} & ResNeXt50 + BERT & 0.9107 & 0.0160 & 0.9359 & 0.9771 & 0.956 & 0.9079 & 0.9731 & 0.9386 \\
          & ResNeXt101 + BERT & 0.9257 & 0.0139 & 0.9458 & 0.9783 & 0.9618 & 0.914 & 0.9711 & 0.9400 \\
          & Vgg16+BERT &\bf 0.9322 & 0.0135 &\bf 0.9496 & 0.9765 & 0.9629 & 0.9217 & 0.9728 & 0.9455 \\
          & Vgg19+BERT & 0.9225 & 0.0158 & 0.9406 & 0.9729 & 0.9564 & 0.9095 & 0.9649 & 0.9343 \\
          & DenseNet121+BERT & 0.9182 & 0.0160 & 0.9379 & 0.9747 & 0.956 & 0.9066 & 0.9698 & 0.9358 \\
          & ResNet50+BERT & 0.9311 &\bf 0.0130 & 0.9487 &\bf 0.9801 & \bf 0.9641 &\bf 0.9237 &\bf 0.9760 &\bf 0.9478 \\
          & ResNet152+BERT & 0.9300  & 0.0142 & 0.9457 & 0.9765 & 0.9609 & 0.9152 & 0.9675 & 0.9391 \\
    \hline
    \end{tabular}%
    \vspace{-10pt}
  \label{tab:imge_impact}%
\end{table*}%

\subsection{Evaluation metrics}

We evaluate all methods in terms of example-based and label-based multilabel classification measures.

{\emph{Example-based metrics}} are defined by comparing the ground-truth label set $\yy=[y_{0}, y_{1}, y_2, \dots, y_C]$ to the predicted label set $\hat{\yy}=[\hat{y}_{0}, \hat{y}_{1}, \hat{y}_2, \dots, \hat{y}_C]$ on each test example, and then calculating the mean value across all test datasets. Subset accuracy  is a strict metric that measures the fraction of correctly classified examples and requires an exact match of  the predicted label set and the ground-truth label set. Subset accuracy is defined as 
$\mathrm{subset ~ accu.} = \frac{1}{N}\sum_{i=1}^{N}\mathbb{I}[\yy_i = \hat{\yy}_i], 
$
where $\mathbb{I}[\yy = \hat{\yy}]$ is an indicator function with value of $1$ when $\yy = \hat{\yy}$, and value of $0$ otherwise. 
Hamming loss evaluates how many labels are incorrectly predicted on average, and is defined as $
\mathrm{hamming~loss} = \frac{1}{N}\sum_{i=1}^{N}\frac{1}{C+1}\sum_{c=0}^{C}\mathbb{I}[y_c \neq \hat{y}_c]. 
$

{\emph{Label-based metrics}} are defined by evaluating the prediction performance of each label separately, and then returning macro- or micro-averaged metric value across all labels.  Precision, recall, and F$_1$-measure are commonly used metrics. F$_1$-measure combines precision and recall, and is the harmonic mean of precision and recall. 
Specifically, for each $c$-th label in $\{0, 1, 2, \dots, C\}$, we denote by $TP_{c}$, $FP_{c}$, $TN_{c}$, $FN_{c}$ the number of true positives, false positives, true negatives, and false negatives, respectively.
Micro-averaged precision, recall, and F$_1$-measure  are defined as follows:
$
 P_{micor}= \frac{\sum_{c=0}^C TP_{c}}{\sum_{c=0}^C TP_{c} + FP_{c}},
R_{micor}= \frac{\sum_{c=0}^C TP_{c}}{\sum_{c=0}^C TP_{c} + FN_{c}}, 
{F_{1}}_{micor}= \frac{\sum_{c=0}^C 2TP_{c}}{\sum_{c=0}^C 2TP_{c} + FP_{c} + FN_{c}}.
$
Macro-averaged precision, recall, and F$_1$-measure  are defined as follows:
$
 P_{macro}= \frac{1}{C+1}\sum_{c=0}^C\frac{TP_{c}}{TP_{c} + FP_{c}},
R_{macro}= \frac{1}{C+1}\sum_{c=0}^C\frac{TP_{c}}{TP_{c} + FN_{c}},
 {F_{1}}_{macro}= \frac{1}{C+1}\sum_{c=0}^C\frac{2TP_{c}}{2TP_{c} + FP_{c} + FN_{c}}.
$
Macro averaging treats all labels equally while micro-averaging favors more frequent labels. High macro-averaged scores usually indicate high performance on less frequent labels, while high micor-averaged scores usually indicate high performance on more frequent labels.

\vspace{-5mm}
\subsection{Baselines}
To the best of our knowledge, our work is the first study on fine-grained detection of illicit drug trafficking on Instagram. We have compared our proposed method with the following baseline methods, which are modified for multilabel learning: 

\begin{packeditemize}
\item Image-only baselines. We fine-tune pretrained CNNs (including VGG16 \cite{simonyan2014very}, ResNet50 and ResNet152 \cite{he2016deep}, and DenseNet121 \cite{huang2017densely}) on the images associated with IDTEs to extract the features for the multilabel classification task. 

\item Text-based baselines. We train a multilabel classification model on the textual information of IDTEs using the following models:  TextRNN \cite{yin2017comparative}, TextCNN \cite{kim-2014-convolutional}, FastText \cite{joulin2016bag}, and BERT \cite{devlin2018bert}.
\item Multimodal learning baselines. We first use ResNet50 and  BERT to extract image and text features respectively, and then apply  multimodal fusion method to combine the image and text features for multilabel classification. We compare following multimodal fusion methods:  concatenation \cite{kiela2018efficient} and factorized bilinear coding (FBC)  \cite{gao2020revisiting,jia20203d}.
\end{packeditemize}

 \vspace{-10pt}
\subsection{Experimental Results} 

Table \ref{tab:classfication}  shows the detection performances of our proposed method  and the baseline approaches with different multilabel classification metrics. Our approach consistently and significantly outperforms the baseline methods on all example-based metrics and label-based metrics. We describe several key observations we have made from these
results as follows:

{\noindent \bf Text vs image in IDTE detection.} We first observe that the performance of text-only methods is better than all image-based methods. For example, even the weakest TextCNN outperforms the best image-based VGG16 by 15.12\% in terms of macro F1.  It indicates that textual information is more important and reliable than images for IDTE detection (contrary to the findings in \cite{yang2017tracking}). We further observe that BERT model can achieve highly accurate classification performance. For example, BERT outperforms the best text-based TextRNN by 37.23\% in terms of macro F1. It demonstrates the superiority of pretrained BERT model in text classification tasks. Finally, it does make a difference when we train image-only models by fine-tuning with different CNNs architectures.
 
\begin{figure}[!t]
  \centering
  \includegraphics[width=\linewidth]{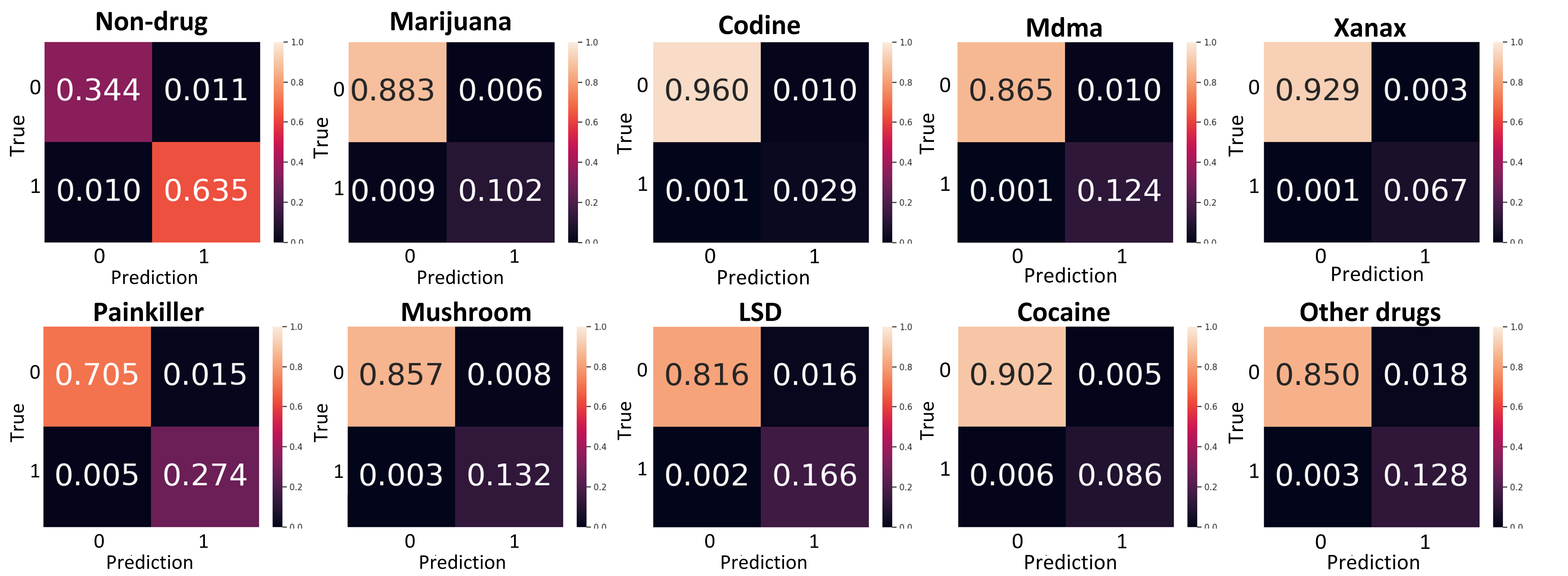}
  \vspace{-20pt}
  \caption{Confusion matrix for each label classified by our method.}
  \Description{}
  \label{fig:confusionMatrix}
  \vspace{-15pt}
\end{figure}

\begin{table*}[htbp]
  \centering
  \small
  \caption{IDTE Detection Case Studies. We compare our method with the best baseline model BERT. Detection errors are highlighted with \textcolor{red}{red} color. We can see that our method delivers more accurate and reliable predictions than BERT.}
  \vspace{-10pt}
    \begin{tabular}{|c|r|p{6.em}|p{15.57em}|p{6.em}|p{15.57em}|}
    \toprule
    \multirow{3}[6]{*}{1} & \multirow{3}*{\includegraphics[scale=0.13]{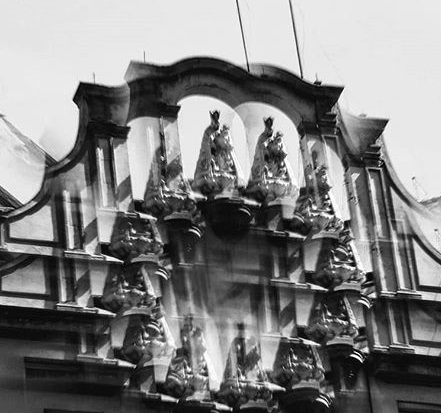}}  & \multicolumn{4}{p{45.355em}|}{\#anxiety\#pain\#depression\#weightloss\#xanax\#oxy\#ritalin\#crystal\#fastdeals\#bitcoinpayment\#trackingnumber \#bestrates. Contact:Wickr:Ste**** Email:ste****@protonmail.com} \\
\cmidrule{3-6}          &       & Groundtruth & \multicolumn{3}{p{38.785em}|}{Xanax, Painkiller, Other drug } \\
\cmidrule{3-6}          &       & Our Prediction & \textcolor{blue}{Xanax, Painkiller, Other drug} & BERT result & \textcolor{blue}{Xanax, Painkiller, Other drug} \\
    \midrule
    \multirow{3}[6]{*}{2} &  \multirow{3}*{\includegraphics[scale=0.065]{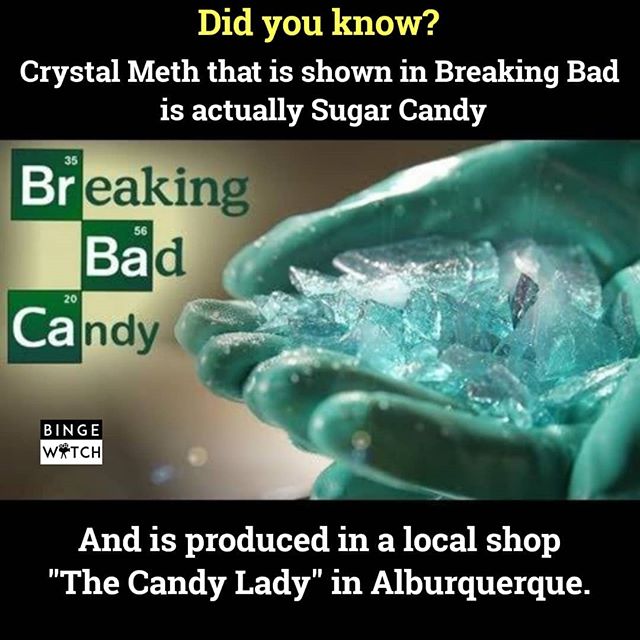}} & \multicolumn{4}{p{45.355em}|}{So \#AaronPaul has been smoking sugar candy from the beginning ?. . . . \#breakingbad \#methamphetamines \#crystalmeth \#brba \#goat \#walterwhite \#bryancranston \#jessepinkman \#got \#drugs \#sugar \#sugarcandy \#thecandylady \#alburquerque \#heisenberg \#danger} \\
\cmidrule{3-6}          &       & Groundtruth & \multicolumn{3}{p{38.785em}|}{Other drugs} \\
\cmidrule{3-6}          &       & Our Prediction & \textcolor{blue}{Other drugs} & BERT result & \textcolor{red}{LSD}, \textcolor{blue}{Other drugs} \\ 
    \midrule
    \multirow{3}[6]{*}{3} & \multirow{3}*{\includegraphics[scale=0.085]{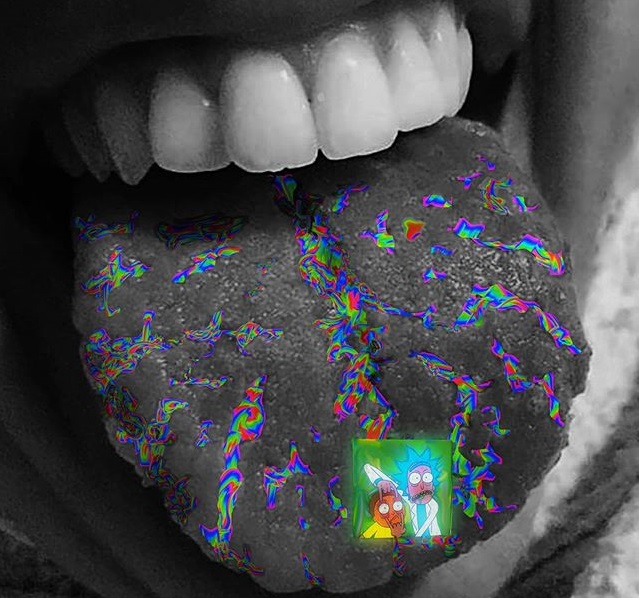}} & \multicolumn{4}{p{45.355em}|}{A.c.i.D, s.H.r.ø.o.M.s..Wickr me:Wins**** .Email:joney****@gmail.com.Whatsapp: (213) ***-3678} \\
\cmidrule{3-6}          &       & Groundtruth & \multicolumn{3}{p{38.785em}|}{Mushroom, LSD} \\
\cmidrule{3-6}          &       & Our Prediction & \textcolor{blue}{LSD} & BERT result & \textcolor{red}{Painkiller}, \textcolor{blue}{LSD} \\
    \midrule
    \multirow{3}[6]{*}{4} & \multirow{3}*{\includegraphics[scale=0.125]{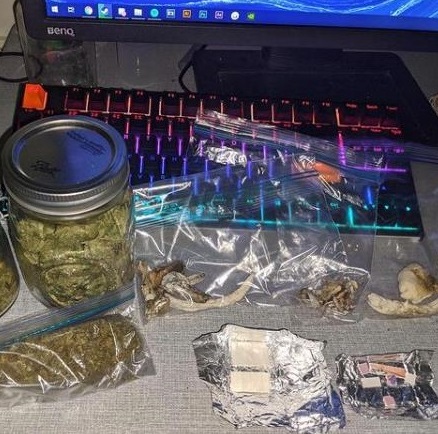}}  & \multicolumn{4}{p{42.355em}|}{$\mathsf{Shr}\square\Box\mathsf{m}\sim\Box\mathsf{C}\Box\mathsf{K}\Box\sim\mathsf{Ac}\Box$d$\Box\sim$Md$\Box$ma$\Box\sim$Esct$\Box$cy$\Box\sim$D$\spadesuit$mt$\Box\sim$w$\Box\Box$d$\Box\sim$Pi$\Box\Box$s$\Box\Box\,\&$mor$\Box\Box\Box.\Box$kik.........speed****7 $\;\Box$wickr...james****7|}\\
\cmidrule{3-6}          &       & Groundtruth & \multicolumn{3}{p{38.785em}|}{Marijuana, Mdma, Painkiller, Mushroom, LSD, Cocaine, Other drugs} \\
          &       & \multicolumn{1}{r|}{} & \multicolumn{3}{r|}{} \\
\cmidrule{3-6}          &       & Our Prediction & \textcolor{blue}{Marijuana}, \textcolor{blue}{LSD}, \textcolor{blue}{Cocaine}, \textcolor{blue}{Other drugs} & BERT result & \textcolor{blue}{Cocaine} \\
    \midrule
    \multirow{3}[6]{*}{5} &\multirow{3}*{\includegraphics[scale=0.29]{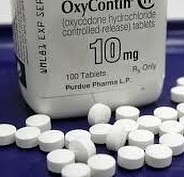}} & \multicolumn{4}{p{45.355em}|}{$\mathsf{WHAT}\boxtimes\mathcal{S}\mathsf{APP}:\,+1(440)\,***-3147.\boxtimes\mathsf{EX}\boxtimes:+1(202)\,\,***-1339.\,\mathsf{WICKR/KIK:}\,$REA****99$\;.\mathsf{SEC}\mathcal{U}\mathsf{RED}/\mathsf{DI}\mathcal{S}\mathsf{C}\mathcal{R}\mathsf{EE}\boxtimes\;\mathsf{DELIVER}\mathcal{Y.}\mathsf{B}\mathcal{UY}......\mathsf{LSD,}\,\mathsf{MDMA,}\,\mathsf{DM}\boxtimes,\mathsf{SHR}\Theta\Theta\mathsf{MS,}\,$ $\mathsf{PILLS,}\,\mathsf{C}\Theta\mathsf{KE,}\,\mathsf{ADD}\mathcal{Y,}$ $\;$XANS,$\,$KE$\boxtimes,\,$2CB,$\,\mathsf{A}\mathcal{N}\mathsf{D,}\,\mathsf{B}\mathcal{U}\mathsf{D}$}\\
\cmidrule{3-6}          &       & Groundtruth & \multicolumn{3}{p{38.785em}|}{Marijuana, Mdma, Xanax, Painkiller, Mushroom, LSD, Cocaine, Other drugs} \\
\cmidrule{3-6}          &       & Our Prediction & \textcolor{red}{Codine}, \textcolor{blue}{Mdma, Xanax, Painkiller, Mushroom, LSD, Other drugs} & BERT result  & \textcolor{red}{Codine}, \textcolor{blue}{Mdma, Xanax, Painkiller, Mushroom, LSD, Other drugs} \\
    \bottomrule
    \end{tabular}%
  \label{tab:use_case}%
  \vspace{-10pt}
\end{table*}%

{\noindent \bf Multimodal learning in IDTE detection.} The motivation of multimodal learning is that the presence of multiple information sources can  be complementary to learn better feature representations for final detection of illicit drugs. However, we observe that multimodal learning does not necessarily outperform models trained with a single modality. As shown in Table \ref{tab:classfication}, two popular multimodal fusion methods (\ie concatenation  and factorized bilinear coding (FBC)) are applied to composite  image and text features extracted from  ResNet50 and BERT respectively; but their performances are  worse then single-modality based BERT model. One possible reason, as discussed in \cite{wang2020makes}, is that such multimodal models are prone to overfitting when features are combined at a late stage. By contrast, building upon a powerful BERT model and projecting image embedding to the text token space, our method achieves better fine-grained multimodal fusion, thus improving the detection performances. For example, our method improves BERT by 2.7\% in terms of macro F1. Figure \ref{fig:confusionMatrix} shows the confusion matrix for each label classified by our model.

{\noindent \bf Impact of image encoders on our method.} 
Thanks to the conceptual simplicity of MMBT, we can easily substitute different image and text encoders into the bitransformer module. Unlike text information, BERT has shown dominating performance; image encoders have shown comparable performance based on Table \ref{tab:classfication}. Therefore, we have conducted an ablation study to compare different image encoders while keeping the BERT encoder the same. Table \ref{tab:imge_impact} includes the performance comparison among seven competing image encoders. It can be observed that on the average ResNet50 achieves the best performance, which is in contrast to the adoption of ResNet152 for MMBT in \cite{kiela2019supervised}.

\subsection{IDTE Detection Case Studies}
The experimental results have shown the effectiveness of the proposed DMML approach in detecting  IDTEs. We  present some case studies as shown  in Table \ref{tab:use_case} to demonstrate the superiority as well as limitations of the proposed method. Case 1 and Case 2 illustrate two examples of the proposed method with completely correct detection, while Cases 3-5 show the examples with missing or false detection. When compared with the results of the text-only BERT model, we can observe that by fusing text with image modality, the proposed DMML can not only effectively reduce false alarms (see Case 2 for example - LSD was incorrectly predicted by BERT but supplementary image information shows it is not LSD;in Case 3, painkiller is false alarm of BERT prediction, which gets corrected by DMML), but also detect more accurate labels (see Case 4 - BERT misses a few labels). However, for input text with special symbols that drug dealers used to disguise illicit deals, the proposed method will fail to accurately detect all  labels, such as the mushroom with `s.H.r.ø.o.M.s' in Case 3, MDMA with `Md$\Box$ma$\Box$' in Case 4, and Cocaine with `$\mathsf{C}\Theta\mathsf{KE}$' in Case 5. One potential solution is to design a new data preprocessing algorithm to recover these special symbols in words.

 \vspace{-10pt}
\section{CONCLUSION}
\label{sec:6}

In this study, we have collected and constructed a large-scale multimodal IDTE dataset (MM-IDTE) from Instagram data to support the research related to illicit drug trafficking activity detection. Our dataset includes both textual and visual information contained in posted comments and has been manually annotated for multiple drug types. An automatic hashtag-based  data crawling system and a user-friendly interactive web-based data annotation system were developed. Our data crawling and annotation systems allow us to build a dataset with thousands of fine-grained samples with multiple labels. Based on the constructed MM-IDTE dataset, we have developed a deep  multimodal  multilabel learning approach to detect suspect IDTEs and demonstrate its effectiveness on the new MM-IDTE dataset. It is shown that the proposed MMBT-based approach can dramatically outperform unimodality and ad hoc multimodal fusion strategies. We have also experimentally compared different image encoders for MMBT-based fusion and found that MMBT based on ResNet50 and BERT achieves the best performance. Extensive experimental results on the MM-IDTE dataset show that the proposed DMML methodology can accurately detect IDTEs even in the presence of special symbols and style changes attempting to evade detection.

\bibliographystyle{ACM-Reference-Format}

\bibliography{refs}

\end{document}